\documentclass[11pt,a4paper]{article}
\usepackage{times,latexsym}
\usepackage{url}
\usepackage[T1]{fontenc}

\usepackage[acceptedWithA]{tacl2021v1}
\usepackage{xspace,mfirstuc,tabulary}

\newif\iftaclinstructions
\taclinstructionsfalse % AUTHORS: do NOT set this to true
\iftaclinstructions

\newcommand{\instr}
\fi

\iftaclpubformat % this "if" is set by the choice of options

\else

\fi

\title{Large Language Models Preserve Semantic Isotopies \\ in Story Continuations}

\title{Large Language Models Preserve Semantic Isotopies \\ in Story Continuations}

\author{
  Marc Cavazza \\
  University of Stirling, UK \\
  National Institute of Informatics (NII), Tokyo, Japan \\
  \texttt{marc.cavazza@gmail.com}
}

\date{}

\usepackage{graphicx}
\usepackage{booktabs}

\begin{document}
\maketitle
\begin{abstract}
 In this work, we explore the relevance of textual semantics to Large Language Models (LLM), extending previous insights into the connection between distributional semantics and structural semantics. We investigate whether LLM-generated texts preserve semantic isotopies. We design a story continuation experiment using 10,000 ROCStories prompts completed by five LLMs. We first validate GPT-4o’s ability to extract isotopies from a linguistic benchmark, then apply it to the generated stories. We then analyze structural (coverage, density, spread) and semantic properties of isotopies to assess how they are affected by completion. Results show that LLM completion within a given token horizon preserve semantic isotopies across multiple properties. 
\end{abstract}

\iftaclpubformat
\fi

\section{Introduction}
The 'unreasonable effectiveness' of Large Language Models (LLM) has fostered a vivid linguistic debate, in particular with respect to LLM semantic theories. The most famous criticism of LLM considers them as devoid of semantics \cite{bender-koller-2020-climbing}. It has received various responses \cite{lappin2024assessing} \cite{sogaard2025language}, which often advocate an internalist \cite{piantadosi2022meaning} or distributionalist position \cite{sahlgren2021singleton}.
Recent work \cite{proietti2025quasi} has revisited the linguistic aspects of LLM from the perspective of distributional semantics \cite{contreras-kallens-christiansen-2025}, following an established position on semantic embeddings \cite{sahlgren2006word} \cite{sahlgren2008distributional} \cite{lenci2018distributional} \cite{lenci2022comparative}\cite{lenci2023distributional}. 
Moreover, the distributional semantics perspective connects to other linguistic theories, in particular structural semantics, via an interpretation of paradigmatic relations \cite{mickus2024language}, while \citet{lenci2023distributional} explicitly mention structuralist linguists such as Hjelmslev and Greimas.
In this paper, we explore the relevance of contemporary structural linguistics models \cite{rastier2015interpretative} to LLM through a dedicated narrative generation experiment, which allows us to investigate aspects of textual semantics \cite{rastier1997meaning} of LLM-generated text. We aim to expand on the original intuition of the above mentioned authors \cite{sahlgren2006word} \cite{mickus2024language}, also taking advantage of the evolution of contemporary structural linguistics towards a textual perspective \cite{rastier1997meaning} \cite{RastierEtAl2002}, which would offer an alternative linguistic framework for LLM semantics. 
Our approach consists in analyzing semantic properties of LLM-generated text using the structural concept of isotopy, a textual semantic representation which will be introduced in the next section. To that effect, we designed a story completion experiment, which consists in extending a short story primer based on an existing dataset (ROCStories \cite{mostafazadeh2016corpus}) to produce a complete story.

\section{Isotopy and Computational Semantics}
Isotopy is a key concept in structural linguistics in that it relates lexical semantics to text and discourse properties. Isotopies can be defined as the repeated co-occurrence of a semantic feature which gives a text its cohesion and guides its interpretation \cite{rastier2015interpretative}. 
For instance, the verses:
\begin{center}
\small % or \footnotesize
There was a fine ship, carved from solid gold \\
With azure reaching masts, on seas unknown
\end{center}
Contain a \emph{navigation} isotopy, occurring through words ‘ship’, ‘masts’ and ‘seas’ which all contain the corresponding semantic feature \cite{hebert2019introduction}. The \emph{navigation} feature (a lexicalized metalinguistic property) is used as the isotopy label.  

\begin{figure}[t]
  \centering
  \vspace{-2mm} % trim space before figure (small, safe)
  \includegraphics[width=\columnwidth]{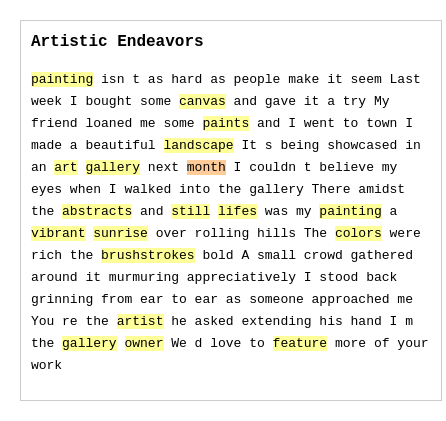}
  \vspace{-1mm} % trim space before caption
  \caption{Isotopy across a story completion. Isotopy constituents are highlighted, and the \emph{pivot} marks the end of the story primer (see text for details).}
  \label{fig:Iso_example}
  \vspace{-2mm} % trim space after caption (optional)
\end{figure}

The interpretative aspect of isotopy is both challenging and essential: it differentiates isotopy from the simple recurrence of related words, as interpretation broadens semantic similarity by resorting to inferential aspects. For instance, isotopies’ constituents may be related by part-whole relationships (‘beach’ and ‘sand’), situational aspects (‘picnic’ and ‘ants’), action and objects (‘write’ and ‘words’); all these examples being drawn from isotopies extracted during our experiments (Figure~\ref{fig:Iso_example}). Although primarily conceived of as a descriptive concept, isotopy has been previously related to advanced concepts in computational linguistics, such as type coercion \cite{peeters2000setting} \cite{willems2013linguistic} in the generative lexicon \cite{pustejovsky-1991-generative}. \citet{attardo2024linguistic} has also resorted to isotopy as an explanatory mechanism for humorous expressions. 
In computational linguistics, the construct that most closely resembles isotopy is undoubtedly the \emph{lexical chain}, introduced in \cite{morris1991lexical} who were influenced by the theory of cohesion in \cite{halliday-hasan-1976-cohesion}. 
A lexical chain is essentially a sequence of semantically related words forming a cohesive thread of meaning, for instance: {‘city’, ‘suburbs’, ‘suburban’, ‘residentialness’, ‘community’, ‘neighboorhood’ …} would all form part of the same lexical chain (excerpt taken from \cite{morris1991lexical}). NLP researchers adopted lexical chains in various systems because they gave access to semantic representations without requiring full sentence understanding \cite{barzilay1999using} \cite{xiong2013lexical} \cite{ercan2007using} \cite{hirst1998lexical}. Various limitations were subsequently identified arising from over-reliance on lexical resources \cite{marathe2010lexical} \cite{galley2003improving} \cite{silber2002efficiently}. However, it is the development of modern embedding techniques which has redefined approaches to lexical chains extraction. With their evolution towards resource-free approaches \cite{remus2013three}, they became conceptually closer to isotopy, however still lacking their interpretative aspects.

\subsection{Other Related Work}
There are obvious connections between distributional semantics, corpus linguistics and the textual approach to semantics we are advocating here. \citet{uchida2024using} analyzed text generated by older LLM such as GPT-3.5 from a corpus linguistics perspective,  comparing collocations to corpus data.  \citet{reinhart2025llms} used corpus-based grammatical and stylistic dimensions to compare human writing to LLM text generation. However, the above works still relies heavily on vocabulary usage \cite{durward2024evaluating}, and are relatively underspecified in semantic terms, in particular on higher-level aspects such as genre detection, which is still based on simple heuristics rather than intermediate semantic representations \cite{rastier1997meaning}.  
Although the focus of our work is not explicitly on textual cohesion, it is worth noting that many of the concepts discussed have been associated to text cohesion which in textual semantics is based on the construction of isotopies. Moreover, notwithstanding debates on the relationship between cohesion and coherence, in textual semantics the former determines the latter, for instance through association between isotopies and discourse topics \cite{rastier1997meaning}.
In this work, we will only be concerned with the initial step of isotopy construction as a textual semantic representation. For the identification of recurring semantic units, LLM-based inference substitutes itself to previous lexical similarity measures that have been used in text cohesion or coherence studies, several of which having been compared in \cite{lapata2005automatic} \cite{budanitsky2006evaluating}, or even approaches that have incorporated distributional aspects \cite{marathe2010lexical} \cite{aletras2013evaluating}. 
At the other end of the processing pipeline, the quantification of isotopies’ properties can be performed by adapting metrics developed for lexical chains such as span \cite{feng-etal-2009-cognitively} and density \cite{hollingsworth2008lexical} providing a simpler and consistent approach than more complex metrics such as \texttt{CohMetric} \cite{seals-shalin-2023-longform}, \texttt{LongWanjuan} \cite{liu-etal-2024-longwanjuan}, or BERT-based metrics for cohesion \cite{he-etal-2022-evaluating}. 

\begin{figure*}[t]
    \centering
    \includegraphics[width=\textwidth]{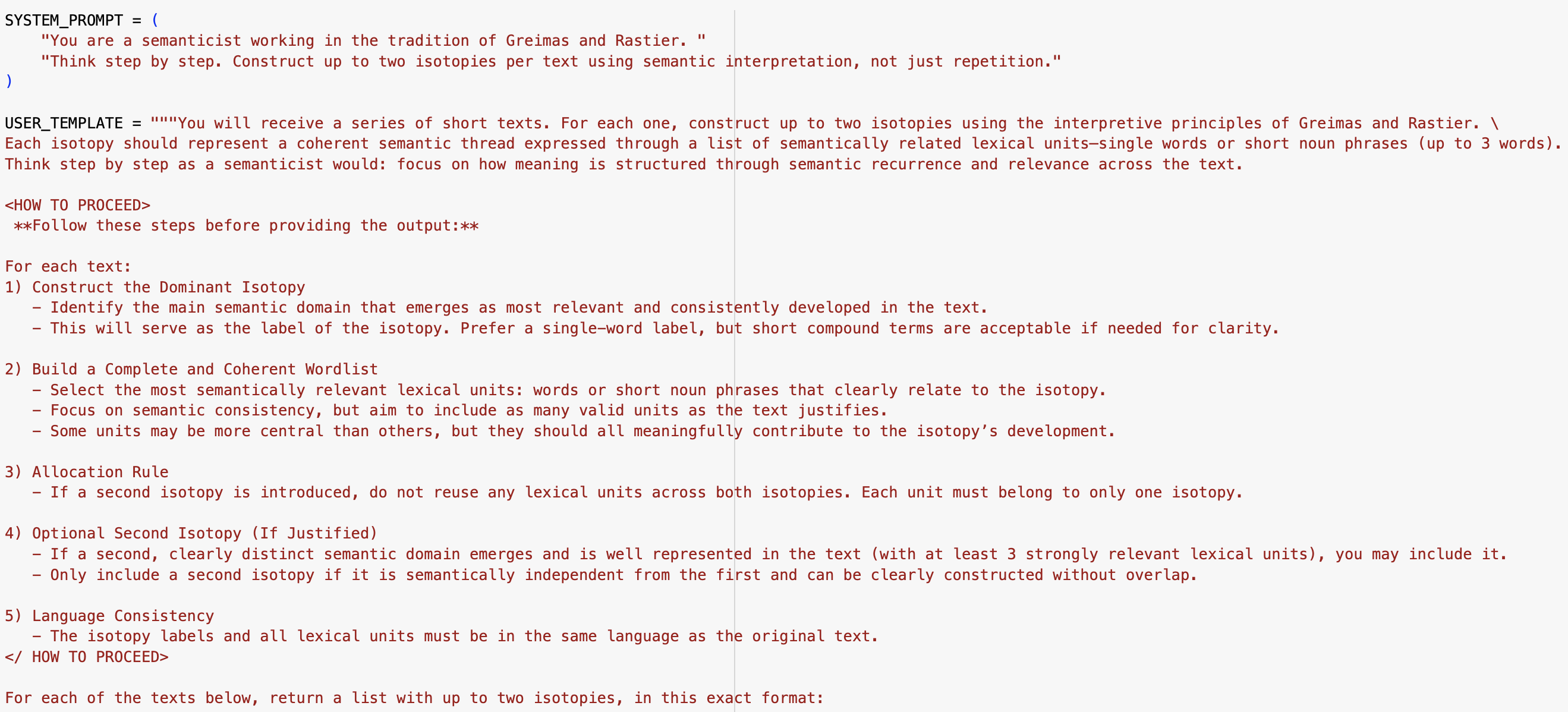}
    \caption{System and User prompt for isotopy extraction (note the various steps inspired from the interpretative approach).}
    \label{fig:prompt_iso}
\end{figure*}

\section{Assessing GPT-4o’s Ability to Extract Isotopies}
In this section, we first establish the ability for LLMs to extract semantic isotopies from texts, in particular GPT-4o which will be used to analyze semantically the story completions provided by the other LLMs. 
In the absence of standard isotopy benchmark, we have collected a set of published isotopies in the specialist literature, which spans across several disciplines in linguistics, poetics and media studies, using the JSTOR\footnote{www.jstor.org} bibliographic database. 

\begin{figure}[t]
  \centering
  \includegraphics[scale=0.50]{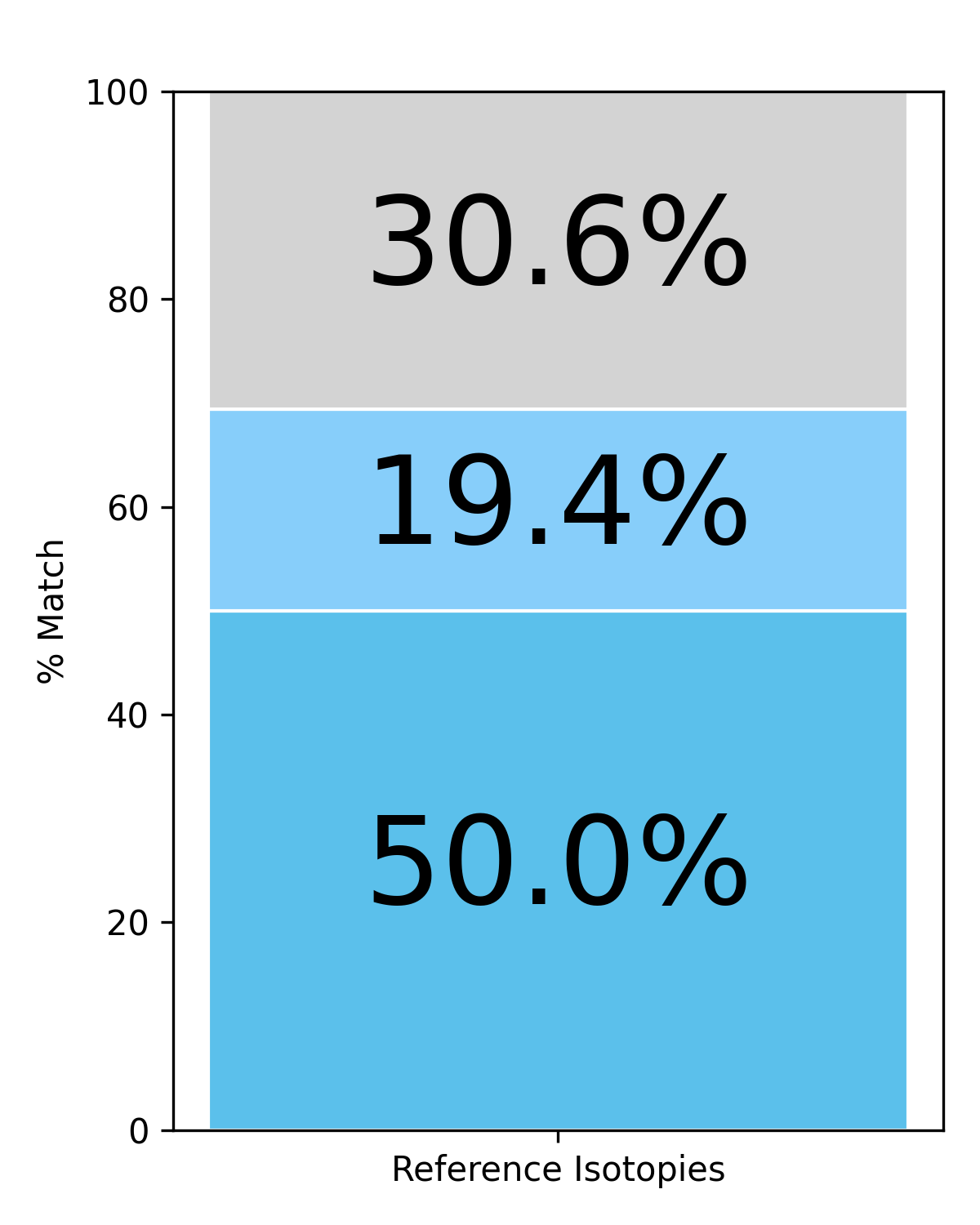}
  \caption{GPT-4o performance on the isotopy benchmark. 50\% of isotopy labels are literal matches, with a further 19.4\% matching with a multilingual embedding.}
  \label{fig:Iso_bench}
\end{figure}

At the end of the process, we have assembled a test set of 26 texts in French (14), Spanish (6) and English (6), corresponding to a total of 36 isotopies. Examples of isotopies include analysis of excerpts from French novels  \cite{rastier1997meaning} \cite{dolores2006piece}, English verses from the Jerusalem Bible \cite{galofaro2020rosa}, or from Shakespeare’s Richard III \cite{ungelenk2023touching}, South American Spanish tales \cite{lins2023analise}, or statements from the head of a central bank \cite{modena2017discours}.
Their underlying texts encompass various styles, while being representative of the challenges faced when extracting isotopies. Our test system relies on running an isotopy-extraction GPT-4o Prompt on the above benchmark, producing up to two isotopies in a Json format, using structured output and temperature $=0$ parameters, with a zero-shot Chain of Thought (CoT) \cite{wei2022chain} instruction, both in the user and system prompt. 

Our quantitative analysis focuses in the first instance on the correct identification by GPT-4o of the isotopy label, which captures how well the LLM has analyzed semantic occurrences, since the isotopy is established via its constituent lexical units, and only subsequently ‘named’. The set of resulting isotopies is then matched to the benchmark, using a comparison procedure that avoids spurious matches when more than two isotopies need to be compared, by exiting upon successful comparison. 
Each isotopy label is first matched literally to the reference and, should no match be found, it is compared via a multilingual semantic embedding (\texttt{XLM-RoBERTa} \cite{conneau-etal-2020-unsupervised}), setting a cosine similarity threshold of $0.70$ for a positive match. Figure~\ref{fig:Iso_bench} reports results, with a literal match level of 50.0\% and an additional 19.4\% embedding similarity, resulting in correct identification in 69.4\% of cases, a conclusive result considering the very challenging nature of the benchmark. We still checked that calculated isotopies contained similar lexical units to those from the reference benchmark, using semantic embeddings. The average cosine similarity between isotopy wordlists across the full test set remains highly significant at $0.76$. Recall based on \emph{literal} matches achieves an average score of 0.59, while using partial matches, in particular for noun phrases, raises this value to $0.77$. 
Overall, considering the highly challenging nature of the benchmark, these results confirm the ability of GPT-4o to extract complex isotopies, and support its use across our story completion dataset.

\section{System Overview}
In this section, we describe our overall processing pipeline for story completion and isotopy extraction from the completed stories. We selected five LLMs for our story completion experiment: 
LLaMA--3.2~3B, Mistral-Nemo~12B, Phi--4~14B, Qwen--2.5~14B, and Gemma--3~27B.

Our objective was to cover various origins (including from a linguistic perspective), training approaches, and model sizes. Of these, three versions are instruction-tuned LLM (Llama--3.2, Qwen--2.5 and Gemma--3) and one has been specifically fine-tuned to storytelling tasks (\texttt{mistral-nemo-storyteller-12B}). We ran preliminary experiments showing that even smaller models succeeded in the linguistic task at hand, \emph{i.e.}, completion with subsequent isotopy extraction. Together with practical experimental constraints, this finding led us to consider mostly average size models. 

We selected the first 10,000 stories from ROC Stories \cite{mostafazadeh2016corpus} and included them in a \texttt{csv} file that was processed by each of the above LLM with a simple completion prompt running locally through LM Studio's server using the Open AI API syntax for all models, which were run with a 4096 tokens window and a temperature setting of $0.5$. 
We opted for the 2016 version of ROC Stories, as it contains titles for each story, which were used at a later stage to explore isotopies’ relevance. 
Each LLM completion file was then analyzed for isotopies using OpenAI's GPT-4o API running online on Google Colab. 
This was implemented using its asynchronous chat-completions API, dispatching batched prompts in parallel, under a bounded semaphore to maximize throughput, while respecting rate limits. Each request minimized decoding variability (temperature = $0$) and was retried up to three times with exponential back-off. We used a bespoke prompt for isotopy extraction which detailed various steps for constructing isotopies within an interpretative framework, with a zero-shot Chain of Thought \cite{wei2022chain}. The prompt did not include an explicit definition of isotopy, nor any examples of isotopies, but mentioned reference theorists (Greimas, Rastier) in the system role prompt (Figure~\ref{fig:prompt_iso}).

\subsection{LLM-based Story Completion}
Completion was driven by a minimalistic prompt, simply instructing each LLM to “continue the following story” with a continuation length of up to 100 words, corresponding to approximately twice the primer length (average: 45 words). No specific instructions were provided as to style, genre, or continuity of topics, not to influence semantic content of completion texts. Across models, the average completion texts were 134 words in length, slightly above the prompt target. However, to properly explore any differences in completion behavior across LLM, we defined a completion ratio as the ratio between the completion length and the primer length (in words).

\begin{figure}[tbp]
  \centering
  \includegraphics[width=\columnwidth]{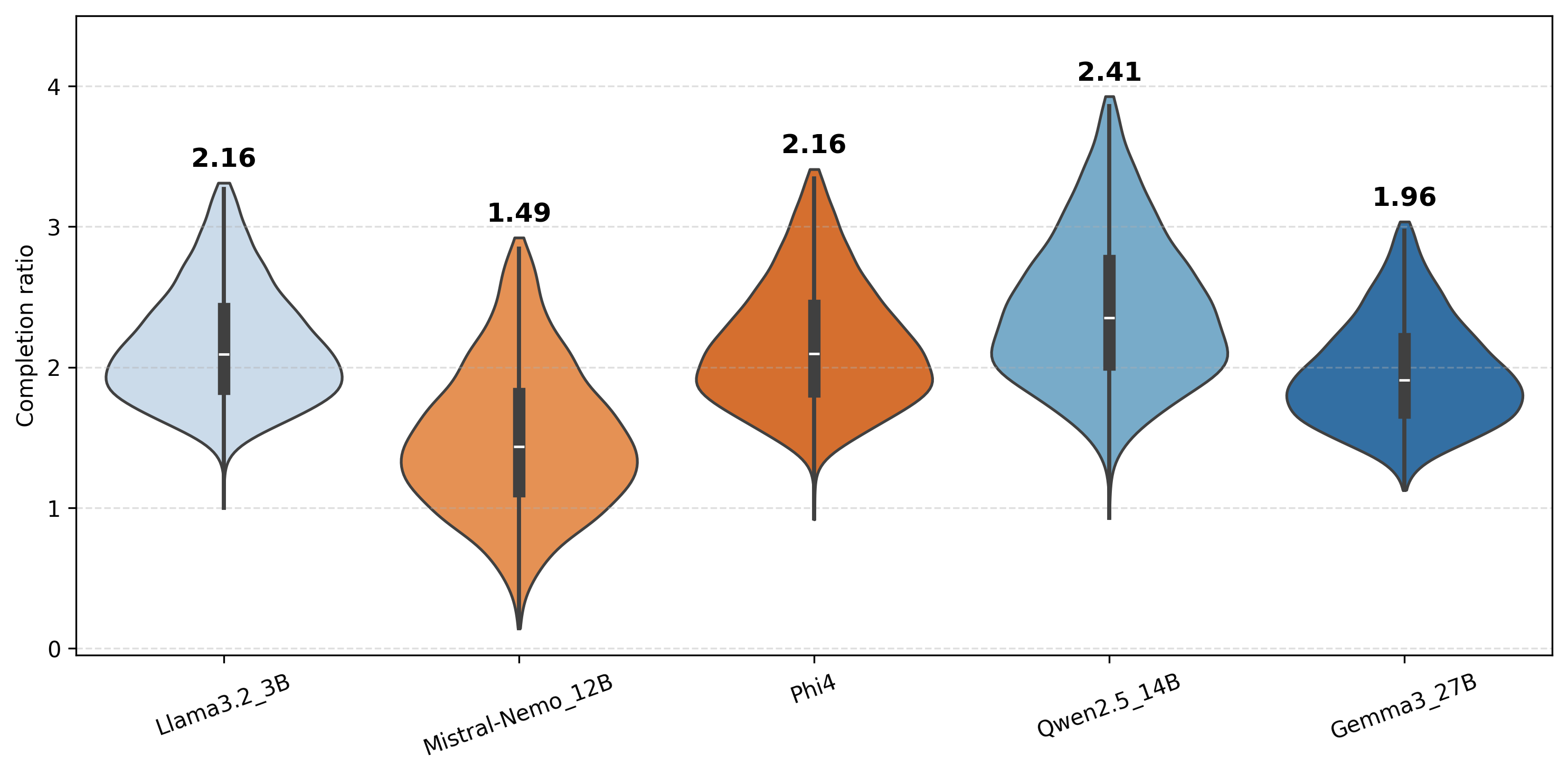}
  \caption{Completion ratio (generated text to original primer) across all five LLM.}
  \label{fig:violinplot_completion_ratio}
\end{figure}

Figure~\ref{fig:violinplot_completion_ratio} represents the distribution of completion ratios for each LLM.
A Kruskal--Wallis test revealed an overall model effect ($H=14156.1$, $p<0.001$, $\varepsilon^{2}=0.290$).
A Dunn's post hoc test (Bonferroni-adjusted) showed nine significant pairwise differences,
with the only large effect sizes observed for Mistral-Nemo 12B (Cliff $\delta$ ranging from $0.55$--$0.79$).
ROCStories primers are comprized of individual short sentences deriving from the initial task they were intended for (Storycloze, or the selection of a story ending sentence \cite{mostafazadeh2016corpus}) with an average sentence length of only $5.9$ words. However, this property did not impact the style of completion text, for which the average sentence length across LLM is $14.6$ words. Post-completion average sentence length varied across models from $10.9$ (Gemma--3 27B) to $17.7$ (Qwen--2.5 14B), a Kruskal--Wallis test showing significant variations ($H = 22503$, $p < 0.001$), with Dunn post hoc (Holm-adjusted) evidencing some large effect sizes, for instance Qwen--2.5~14B vs.\ Gemma--3~27B ($\delta = +0.914$).

\subsection{Extracting Isotopies from Completed Stories}
Our default prompt extracts up to two isotopies to account for borderline cases. However a second isotopy was generated for only 11.4\% of completed texts. In view of this imbalance, we decided to only retain one isotopy per text (the longest one); this resulted in 10,000 isotopies for each LLM completion run Figure~\ref{fig:Iso_example}. Since the objective of our experiment is to explore how isotopies are preserved by LLM completion, we needed to remove ill-formed isotopies, \emph{i.e.}, any isotopy that fails to cover the primer, as no continuation could be computed in that case. Overall, such anomalies account for 2.58\% of all isotopies, bringing the total of well-formed isotopies to 48710. Example isotopies are shown on Table~\ref{tab:isotopies}.

% --- Toggle caption on/off ---
\newif\ifshowisocaption
\showisocaptiontrue      % set to \showisocaptionfalse to remove the title

\begin{table}
\centering
\small                      % slightly smaller font (TACL-friendly)
\setlength{\tabcolsep}{6pt} % modest cell padding
\renewcommand{\arraystretch}{1.1} % a bit more line height
\begin{tabular}{p{0.9\columnwidth}}
\toprule
\textbf{Isotopies} \\
\midrule
\textit{Politics}: \{president, debate, inauguration, oath, executive order\} \\
\textit{Hydration}: \{drink, water, gallon jug, filled, water intake, drinking\} \\
\textit{Therapy}: \{therapist, empathy, clients, anxieties, self-discovery, patients, consultation, helping\} \\
\textit{Car accident}: \{driving, stop sign, crashed, van, alarms, seat belts, damage, distracted driving\} \\
\textit{Injury}: \{cut, hospital, medical aid, doctor, stitching, infection, bandage, accidents, minor setback\} \\
\bottomrule
\end{tabular}
\ifshowisocaption
\caption{Examples isotopies extracted from completed stories, with their constituent lexical units (16.8\% of all lexical units contain more than one word).}
\label{tab:isotopies}
\fi
\end{table}

\section{Analyzing Isotopies}

To constitute a meaningful semantic representation, it is important for isotopies to extend through a sufficiently large fraction of the text. By analogy with lexical chains, for which span was defined as the text covered by the chain \cite{barzilay1999using} or the distance between the first and the last entity in the chain \cite{feng-etal-2009-cognitively}, we define coverage as the distance between the first and the last element of an isotopy, divided by the segment length. A coverage score can be associated to each isotopy relative to the text from which it is extracted, with a maximum coverage score of $1.0$ corresponding to an isotopy spanning across the entire text.

\begin{figure} [b]
  \centering
  \includegraphics[width=\columnwidth]{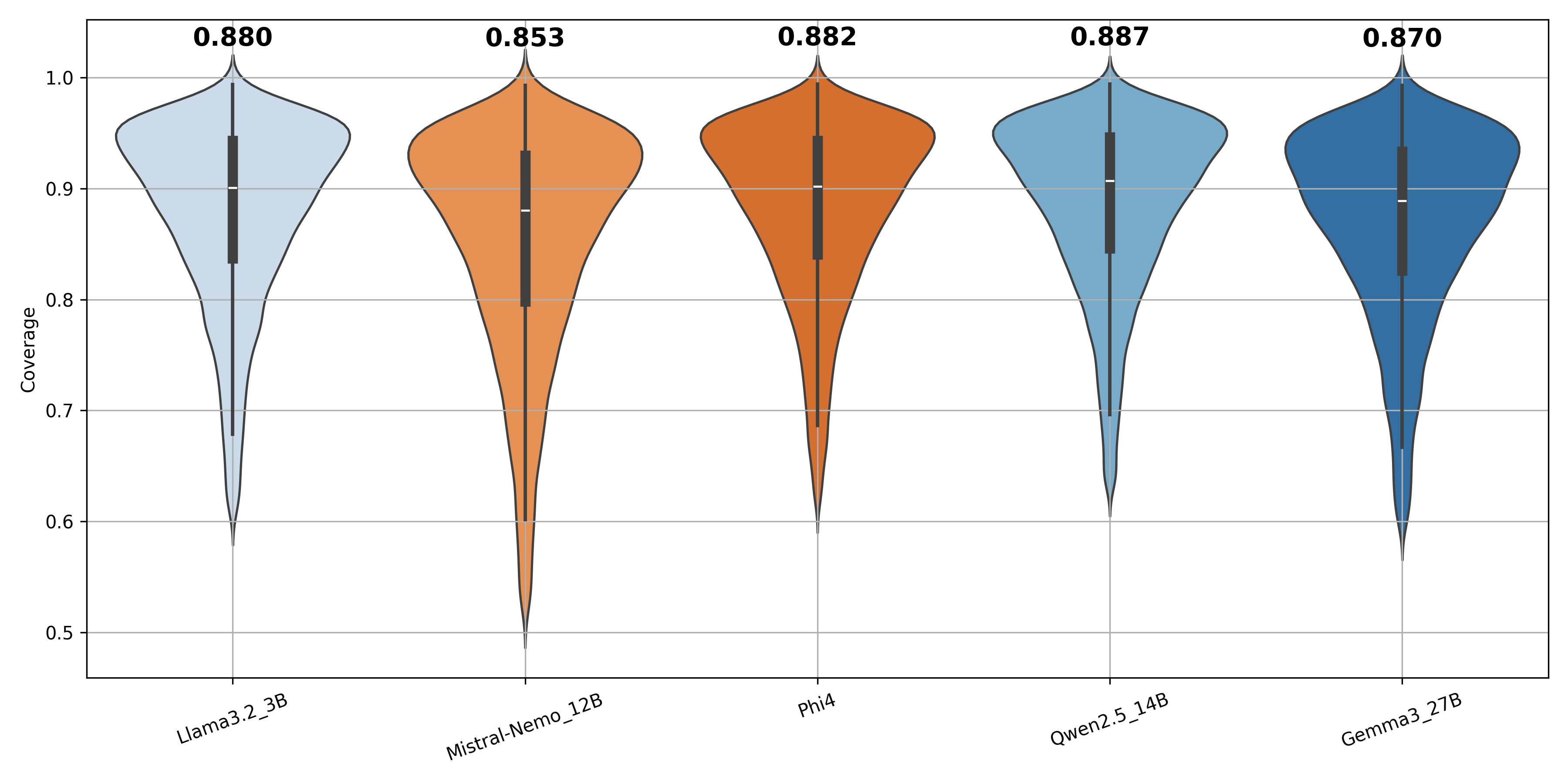}
  \caption{Global Coverage for isotopies extracted for each LLM--generated Completion. All models achieve significant coverage of the text.}
  \label{fig:Coverage_global}
\end{figure}

We compare the distributions of coverage for all five LLM:  the plots of Figure~\ref{fig:Coverage_global} confirm that the isotopies cover a substantial portion the completed text, with all their average coverage scores approaching 1. 
Note that for plot readability a trimming of long-tail extreme values (IQR with Tukey $k = 2$) was performed; 
this resulted in the removal of $2.5\%$ of data points from the original set of $48{,}710$ well-formed isotopies. 
A Kruskal--Wallis test confirmed differences across models ($H = 663.03$, $p < 0.001$), 
which was followed by a Dunn post hoc (Holm-adjusted) with Cliff $\delta$ effect sizes. 
Only two Cliff $\delta$ show some small effect size, both involving Mistral--Nemo (against Phi--4, $\delta = -0.156$, and against Qwen~2.5~14B, $\delta = -0.190$).

\subsection{Coverage Balance}
\begin{figure}[tbp]
  \centering
  \makebox[\columnwidth][l]{% left-align image box inside column width
    \hspace*{-0.5em}% small left shift, adjust as needed
    \includegraphics[
      width=1.05\columnwidth, % slightly larger than the column
      height=0.47\textheight, % taller but still proportional
      keepaspectratio
    ]{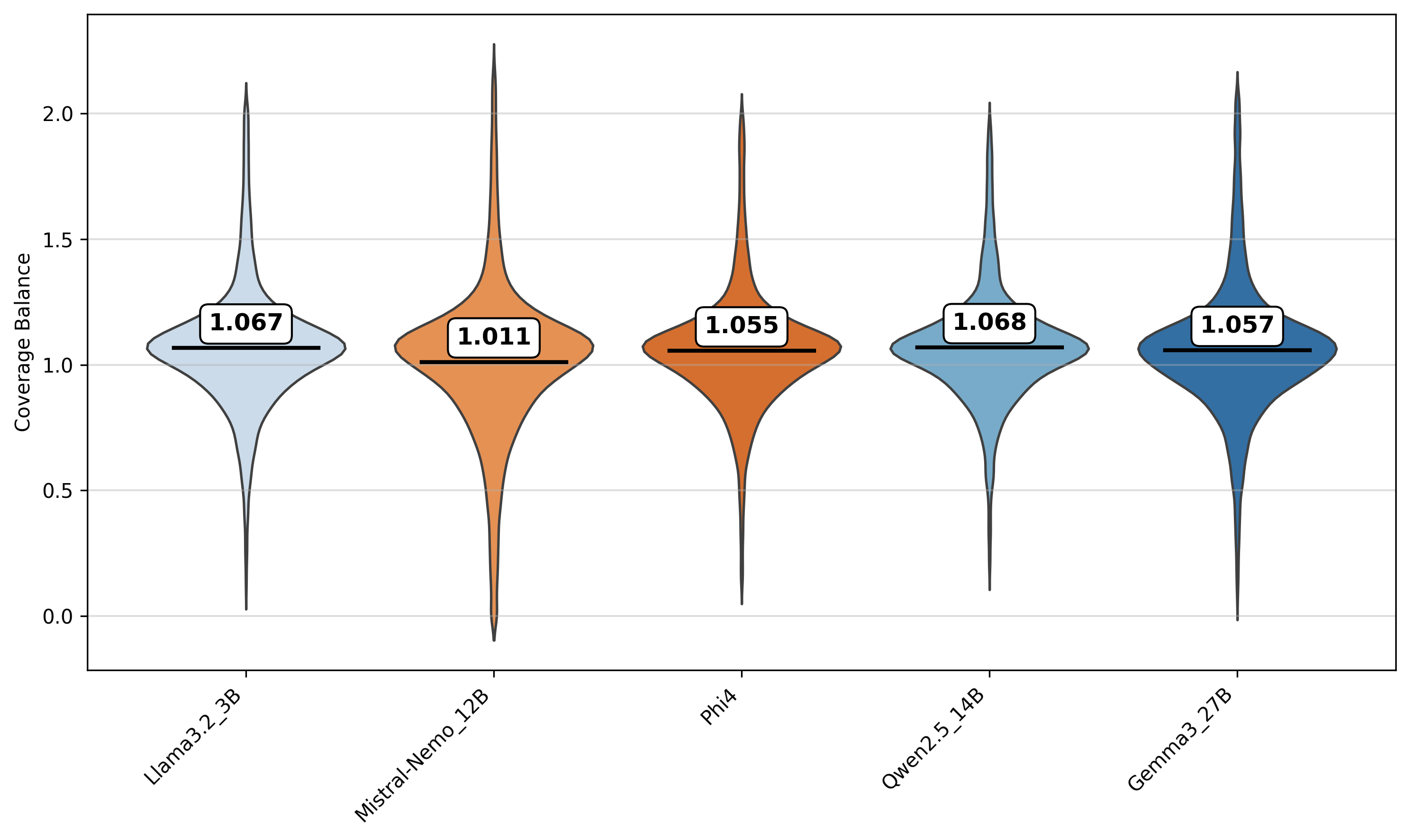}
  }
  \caption{Coverage Balance for each LLM. Labels show mean values, all of which are close to the \emph{balanced} score of $1.0$}
  \label{fig:coverage_balance}
\end{figure}

Our main research question considers whether LLM completion preserves isotopy over the whole completed story, as this would be an indication that LLM have the ability to capture linguistic properties at the textual level. This property needs to be quantified to support a more detailed analysis. To that effect, we introduce one simple metric, which measures how the main isotopy is balanced across primer and completion. 
Now that we have described coverage for an isotopy, we can then formulate coverage balance as the ratio of post-completion to pre-completion coverage: a value close to $1.0$ indicates perfect balance, hence the appropriate continuation of the isotopy from the primer to the LLM-generated completion. 
We compute coverage balance across each individual isotopy obtained for each LLM completion, with the aim of comparing distributions as well as average values across LLM. Following an exploratory analysis, we uncovered significant outliers that compromised plot readability and interpretation. We therefore removed them using the interquartile range (IQR) method with multiplier 
$k=4$, retaining 45,889 isotopies across five models from the original 48,710 (a reduction of 5.79\%). This level of reduction, despite a conservative IQR Tukey coefficient ($k=4$) should be attributed to the long-tailed distribution of coverage.
Figure~\ref{fig:coverage_balance} plots the distribution of coverage scores across the five LLM as a violin plot. In this figure, we rank the models left to right, from smallest (3B) to largest (27B) and adopt the following color code: ‘instruct type’ LLM in blue, and other models in orange, with darker shades corresponding to larger models. 

The average coverage balance across all five LLM is close to the optimal figure of $1.0$, with marginal differences between models that still warrant a proper statistical analysis. A Kruskal--Wallis test confirms differences in the sample with $H = 215.5$, $p$-value $< 0.001$, $\varepsilon^{2} = 0.005$. We ran a post hoc Dunn test (with Holm correction) that evidenced pairwise differences, however with Cliff $\delta$ values ranging from $+0.103$ (Llama--3.3~3B vs.\ Mistral--Nemo~12B) to $-0.109$ (Mistral--Nemo~12B vs.\ Qwen--2.5~14B), still all corresponding to minor effect sizes. 
All LLM completions resulting in balanced isotopies coverage, regardless of LLM size or training type, with a moderate difference in favor of the only narratively specialized model (Mistral--Nemo~12B), which scores best. This first analysis suggests that LLM-generated text indeed preserves semantic isotopy. 

\subsection{Density}
It is challenging, yet important to find characterizations of isotopies that could help determining their consistency with the linguistic descriptions available. The concept of density accounts for which fraction of a given text’s words pertain to the isotopy.  A similar metric had been introduced for lexical chains, although various authors have given it slightly different definitions. The one we are using here is closer to its lexical chain equivalent found in \cite{hollingsworth2008lexical} 

\begin{figure}[b]
  \centering
  \includegraphics[width=\columnwidth]{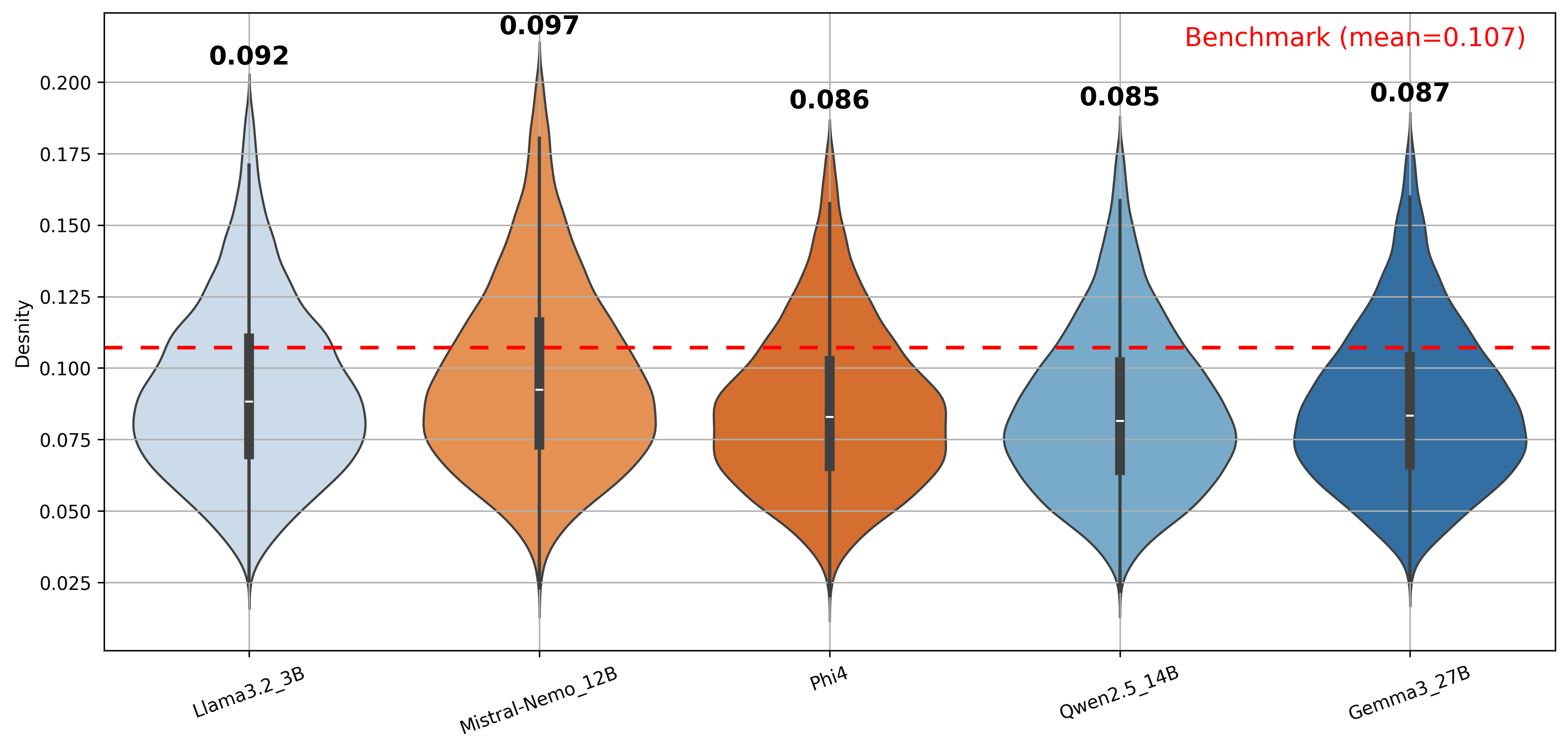}
  \caption{Density of the isotopies extracted from each LLM Completion. Average benchmark texts density is represented as a red dashed line.}
  \label{fig:Density_bench}
\end{figure}

Figure~\ref{fig:Density_bench} shows isotopy density distributions for all five LLM as well as their average values. The red dotted line represents the average density value for texts in our isotopy benchmark, whose average length is however smaller than that of our completed stories\footnote{It should be noted that density tends to increase for shorter samples, which could explain the higher density observed for our benchmark texts: $F(1,36) = 22.50$, $p < 0.001$, $R^{2} = 0.385$, $f^{2} \approx 0.63$}.   
A Kruskal--Wallis test ($H = 841.4$, $p < 0.001$) confirms that density differs across models. 
However, a post hoc Dunn test (Holm corrected) suggests that only a few pairwise comparisons achieve even a small effect size (Mistral--Nemo vs.\ Qwen--2.5~14B, $\delta = 0.21$; Mistral--Nemo vs.\ Phi--4, $\delta = 0.19$; Mistral--Nemo~12B vs.\ Gemma--3~27B, $\delta = 0.175$).
We also compared the density average values for each LLM to the average density of the benchmark, despite the huge difference in sample size, as it constitutes our only available ground truth. We used a Mann--Whitney $U$ test (again with Holm correction to control the family-wise error rate at $\alpha = 0.05$): after adjusting for multiple tests, none of the observed differences reached significance levels. 
Overall, this analysis suggests that isotopy density over completed texts is congruent with human-extracted isotopies on human-authored texts, confirming that LLM completion led to isotopies showing realistic properties on this additional metric. 

\subsection{Isotopy Spread}
After considering isotopies’ coverage and density, which are global metrics, we need to find a characterization of how uniformly their constituent words are spread throughout the text. 

\begin{figure}[t]
  \centering
  \includegraphics[width=0.85\columnwidth]{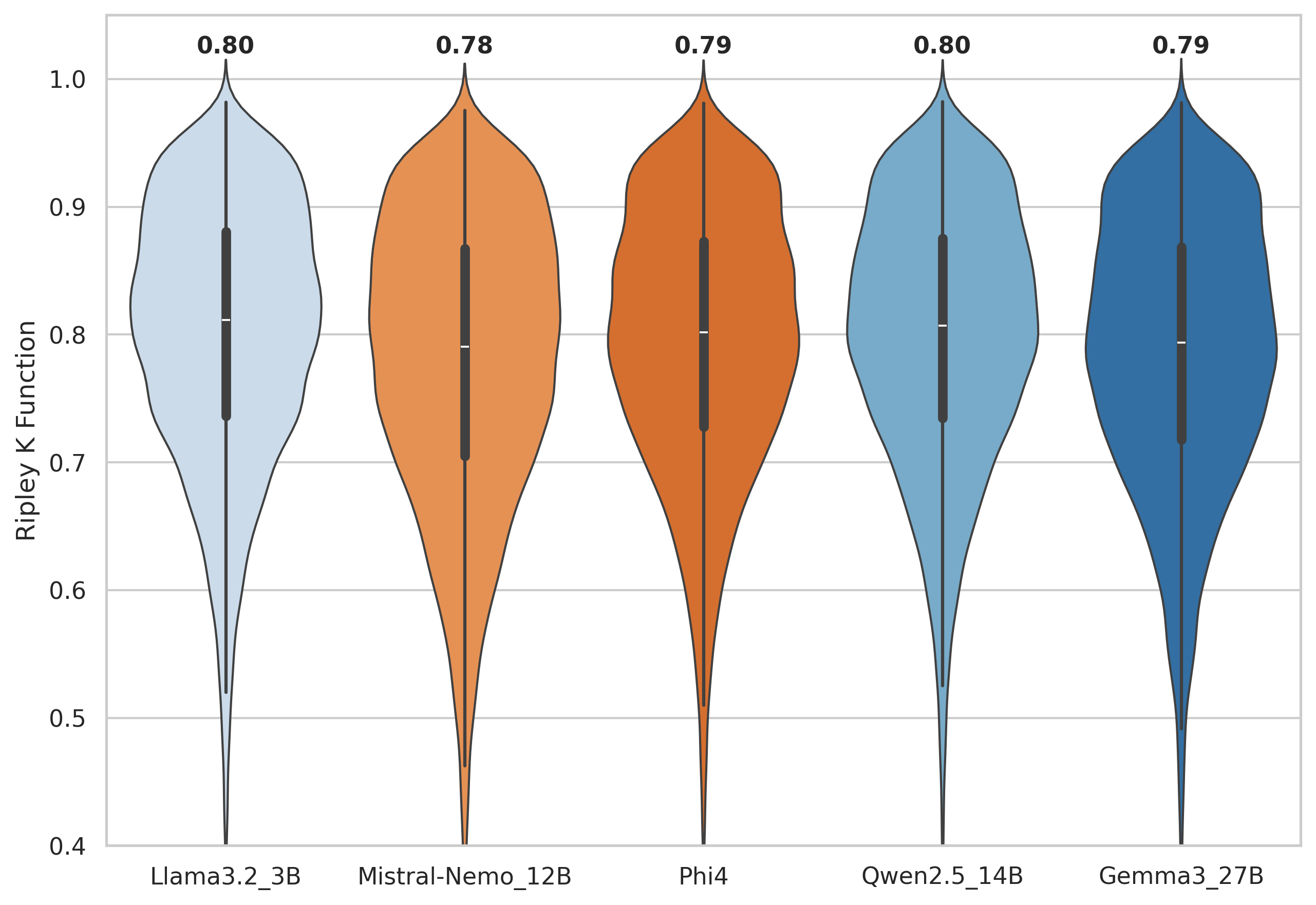}
  \caption{Distributions of Ripley's K functions indicating how isotopies spread across the text (a value of $1.0$ indicates perfect spread).}
  \label{fig:Ripley_violin}
\end{figure}

Among several techniques that could be applied to measure homogeneity of isotopy words distributions (including entropy-based ones), we have selected the one-dimensional version of Ripley’s K score \cite{ripley1977modelling} for its emphasis on spatial aspects. Across the five LLMs, Ripley's $K$ score varies between $0.78$ and $0.80$, out of a maximum normalized score of $1.0$, which would correspond to a perfectly even spread of isotopy across the text (Figure~\ref{fig:Ripley_violin}). A statistical analysis comparing LLMs against each other suggests variations across models (Kruskal--Wallis, $H = 240$, $p < 0.001$), however post hoc analysis with a Dunn test (Holm-corrected) shows limited effect size (maximum Cliff $\delta = 0.109$).
Overall, these results are in favor of a near-even spread of isotopy words throughout the completed text.

\section{Isotopies' Semantic Properties}
The above results indicate a stability of isotopies’ structural properties across completed stories, suggesting that isotopies extracted over completed text retain similar properties to natural isotopies. Because we have concentrated on one single isotopy for each text, a further element of validation is to consider the actual semantic properties of the isotopy.

\begin{figure}[b]
  \centering
  \includegraphics[width=\columnwidth]{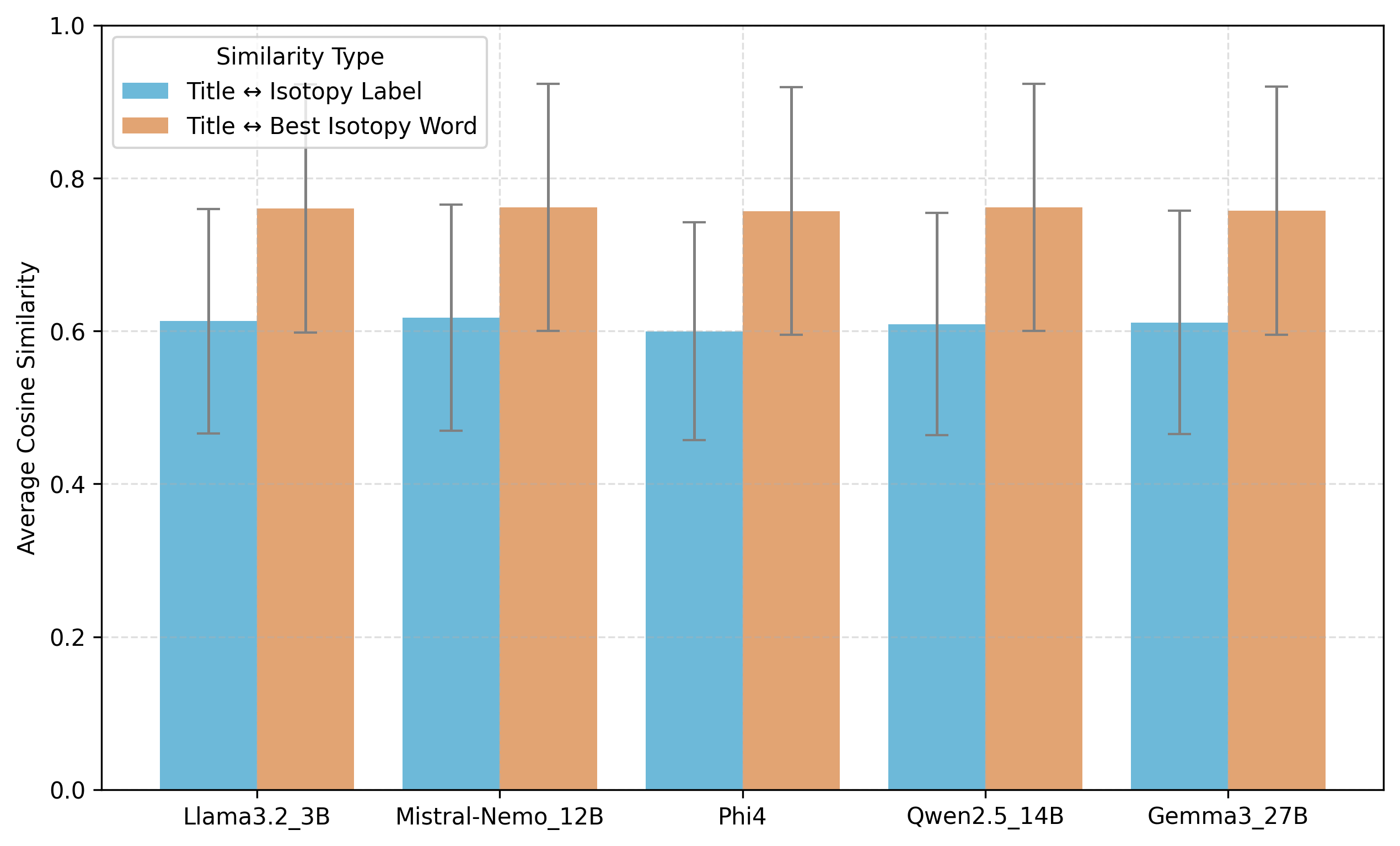}
  \caption{Measuring similarity between story titles and isotopies.}
  \label{fig:Similarity_histogram}
\end{figure}

\subsection{Isotopies' Semantic Relevance}

Any direct semantic comparison between the isotopy label and the text itself might be circular in view of the very definition of an isotopy, and with the total number of isotopies (40,000+) any manual validation would be prohibitive, notwithstanding the scarcity of experts or the limitations of other approaches, such as crowdsourcing, in this context. 
However, each story in ROCStories  is associated a title, which can be used as a proxy for the story main topic, to determine the isotopy's actual relevance. This might not always constitute a ground truth, as some titles may be figurative: for instance, the title ‘foolish Frank’ corresponds to a story where the hero gets drunk and whose main isotopy is about alcohol. 
In an effort to determine whether the main isotopy extracted can be deemed to properly characterize the main story topic, we have devised a semantic matching method. We use semantic embeddings to measure the similarity of story titles (average length 2.2 words) to isotopy labels. We have used \texttt{mxbai-embed-large-v1} \cite{lee2024open} as an embedding model, as we are only dealing with English vocabulary. In addition, we are matching the story title to each isotopy constituent units and selecting the highest cosine similarity. Results are presented on Figure~\ref{fig:Similarity_histogram}, again for each model. The average cosine similarity between titles and isotopy labels stands at $0.6$, which is indicative of relevance without reaching the threshold for high similarity. However, this is actually the case when considering the best isotopy unit match, with an average cosine similarity of $0.8$. 

\subsection{Isotopies' Labels Distributions}
\begin{figure*}[t]
    \centering
    \includegraphics[width=\textwidth]{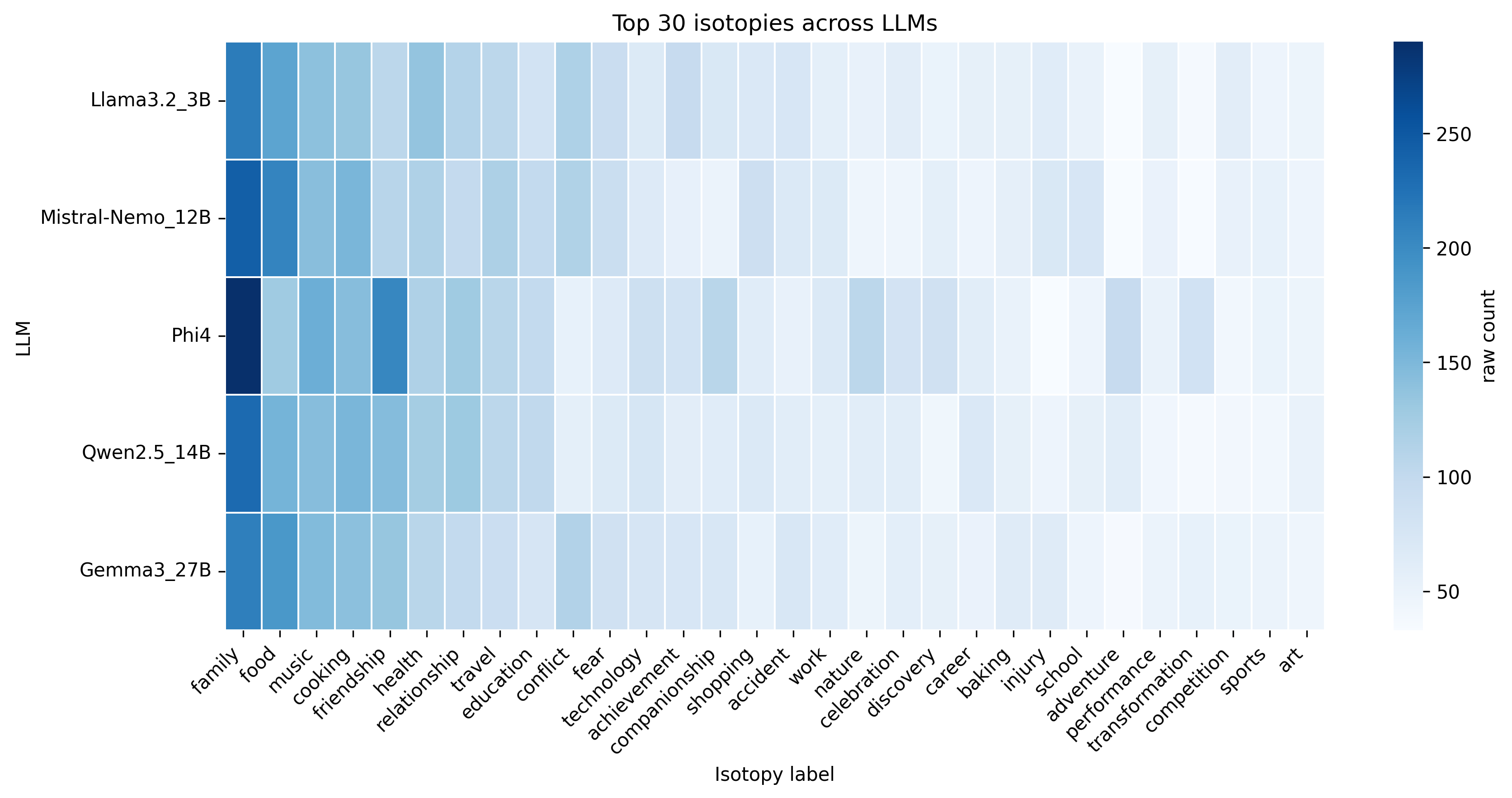}
    \caption{Heatmap for the top 30 isotopies across LLMs, showing convergence of topics and LLM variations. Isotopies' labels convergence does not affect lexical variabilty in isotopies (see text for discussion).}
    \label{fig:top_isotopies_heatmap}
\end{figure*}

The dataset we have used, ROCStories \cite{mostafazadeh2016corpus}, is heavily oriented towards everyday life narratives. As such, one may expect a certain recurrence of narrative topics, with the consequence that isotopies will exhibit a similar convergence in view of their role in identifying narrative topics \cite{rastier1997meaning}. We investigated whether this was the case, and whether there were any major discrepancies between the various LLM that could be explained by model size or training type, in particular when this could impact the creative aspects of text generation.
We found a strong convergence across models with many texts described through similar isotopies (corresponding to the story main topic), although this has no straightforward implications on the actual words constituting these isotopies, which only depend on the text’s vocabulary. Figure~\ref{fig:top_isotopies_heatmap} represents a heatmap for the top 30 isotopies (which are repeated on average 429 times each, and account for 26\% of all isotopies). 

\begin{figure}[b]
  \centering
  \includegraphics[width=\columnwidth]{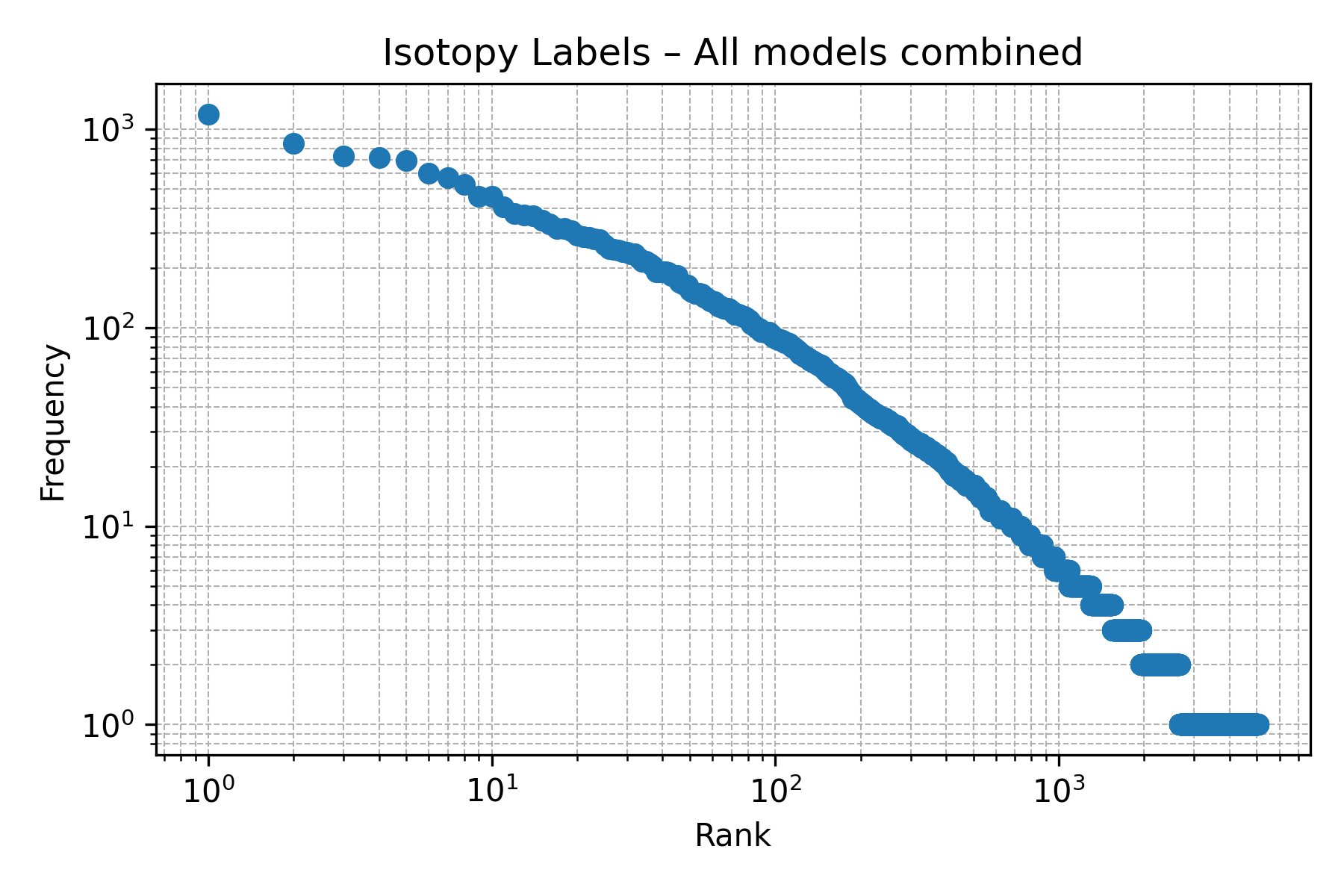}
  \caption{Rank–Frequency Distribution of Isotopies' Labels Across Models.}
  \label{fig:zipf_all_models}
\end{figure}

The top 500 isotopy labels account for 74.5\% of the total occurrences. To compare labels rankings across all models, we applied a Kendall W test to the top 500 isotopy labels, resulting in a Kendall W = 0.814, p < 0.001, which indicates a strong agreement across LLM on the top isotopies, covering a substantial fraction of total occurrences. Pairwise comparisons between individual models show Kendall's $\tau$ scores varying between $0.55$ (Mistral-Nemo~12B vs. Phi--4) and $0.68$ (Llama--3.2~3B vs. Gemma--3~27B). Isotopy repetition leaves an average of $2320$ different isotopy labels for each LLM out of an average of $9742$ occurrences (i.e., after removal of ill-formed isotopies). Figure~\ref{fig:zipf_all_models} suggests a Zipfian \citep{zipf1949human} distribution for the isotopy labels, aggregated across all models (individual models show a similar pattern with minor variations, not represented here). When all models' isotopy labels are pooled, the combined curve has a slope of $-0.82$ and its long tail ($\leq 5$ mentions) now accounts for $78.4\%$ of labels, showing how diversity accumulates across models.
While it is true that ROCStories converge on a limited number of thematic domains, our results indicate that isotopy preservation cannot be reduced to trivial topic overlap. First, isotopy labels are consistently reused across models, but their lexical realizations differ substantially, revealing model-specific strategies for instantiating cohesion. All top 30 isotopies  display strong vocabulary divergence (diversity index $> 0.5$), and none exhibit a stable shared core (core coverage $< 0.5$). 
For instance, the isotopy ‘cooking’ occurs 723 times, with an average length of 8.1 words per occurrence, yet 1215 distinct lexical units overall.
Second, the long-tailed Zipfian distribution of isotopy labels shows diversity beyond a small set of repeated topics. Finally, many isotopies rely on inferential relations rather than lexical overlap, confirming that LLMs sustain for isotopy generation on a linguistic basis rather than merely aligning on story topics.

\subsection{Interpretative Aspects}
The extraction of isotopies follows interpretative semantics principles implying that constituent words from an isotopy might be connected via inference rather than direct, static, lexical relations; this is an integral part of the linguistic framework and is reflected in our benchmark, but from a computational perspective it was largely inaccessible before the advent of LLMs. It is implemented through detailed prompting instructions (Figure~\ref{fig:prompt_iso}). 
To assess that isotopies are formed beyond straightforward lexical relations between their component units, we examine the connections among an isotopy’s elements using standard lexical resources like WordNet \citep{Fellbaum1998}. A strong proportion of standard lexical relations between an isotopy’s words would argue against an interpretative strategy.
To determine the global relatedness of isotopy units from a WordNet perspective, we compute similarity for each word pair in the isotopy, which is then averaged over the whole isotopy. 
To compute similarity, we take the Cartesian product of the words' relevant \texttt{synsets}, apply NLTK's \texttt{path\_similarity} function, and retain the highest score. 
The \texttt{path\_similarity} function operates by finding the minimal number of edges connecting the two \texttt{synsets} across hypernym/hyponym links (moving up and down in the taxonomy).
This score characterizes semantic proximity based on direct lexical relations. 
Applying it over the full set of isotopies, $85\%$ of isotopies score below the weak-link threshold of $0.20$, with only $1\%$ reaching the $0.40$--$0.60$ similarity band. 
Despite the limitations of this sort of similarity computation (in terms of \texttt{synset} choice), this constitutes a heuristic confirmation of the contextual nature of semantic relations within isotopy lexical units. 
Because \texttt{path\_similarity} is purely taxonomic, it ignores many relations determined by roles, attributes, or common-sense knowledge (\emph{e.g.}, cause--effect) that require inference and are part of many isotopies extracted in our experiment.
These aspects can be illustrated by two examples from our dataset. 
The following \emph{perfume} isotopy, \{perfume, smell, scent, fragrant, olfactory\}, achieves a high \texttt{path\_similarity} score of $0.83$, corresponding to direct lexical relations between its constituents. 
Conversely, this \emph{allergy} isotopy, \{itch, allergic, reaction, hives, Benadryl, symptoms\}, only achieves a score of $0.11$ and is clearly predicated on contextual knowledge (in a sense, here from symptom to diagnosis to treatment). 
Note that in this case there is not even a multi-word issue for \texttt{synset} extraction, since \emph{allergic reaction} is recognized as two separate lexical entries.

\section{Discussion}
Our work has also a number of limitations, primarily related to the nature and scope of the experiment itself. Although we have produced evidence supporting the continuation of isotopies through story completion, current LLM literature suggests a loss of stability beyond a generation horizon of 4000 tokens \cite{liu-etal-2024-longgenbench} \cite{bai2025longwriter}. In view of the size of the texts we have studied and generated, which are well within this limit, we should be cautious about generalizing our findings too early, as these will need to be replicated on longer generated texts. 
In addition, we have focused our work on story completion only, which could have influenced linguistic properties captured through isotopy. However, we have gathered anecdotal evidence of isotopy preservation on non-narrative texts, for instance completion for prompts extracted from the Argilla database\footnote{https://huggingface.co/datasets/argilla/prompt-collective}. 
ROCStories may not be fully representative of human narrative style, owing to their predominantly short-sentence composition. However, the resulting isotopy metrics on completed texts remain consistent with observations on benchmark texts, which is in favor of the completion producing realistic texts, and the convergence around some topics did not affect lexical variation in extracted isotopies. 

Previous work has suggested that some instruction tuned models \cite{yuan-chen-ng-2025-instruction} have demonstrated marked gains in narrative coherence in story generation. Notwithstanding the complex relations between coherence and cohesion, isotopy being closer to the latter, we have not observed similar effects.
Isolating the direct effects of instruction-based fine-tuning on story completion and subsequent isotopy extraction remains challenging. Throughout this study, we have uncovered no major difference between instruct-based LLMs and others, in any of the above analyses, although this could be attributed to our instruction models having followed generic tuning only. 
In a similar fashion, we found no substantial impact of model size on performance; the uniform parameterization of all token windows to 4k, as well as the limited length of our completed texts may have evened out some potential differences across models. 
Our only recurrent finding has been that the one model specifically tuned towards storytelling (Mistral-Nemo 12B) has exhibited differences in behavior related to completion length, sentence length, and most of the isotopy metrics, albeit not always reaching notable effect sizes. 

\section{Conclusions}
We have provided initial evidence that LLM-based generation preserves textual semantics properties. By expanding on intuitions from previous work \cite{sahlgren2006word} \cite{lenci2023distributional} \cite{mickus2024language} we elaborated further on the relevance of structural semantics, extending its description to the textual level.
At this stage, it is difficult to hypothesize how such textual properties have been captured during LLM corpus training, and which mechanisms support them. It is generally considered that nucleus sampling \cite{holtzman-etal-2020-curious}, by reducing degenerate outputs may contribute to improve cohesion, although its reduction in repetitive outputs may seem to contradict some basic requirements of isotopy, notwithstanding qualitative aspects. \citet{meister-etal-2023-efficacy} empirically demonstrate that sampling adapters enhance precision in generation and improve sequence-level quality, as measured by metrics such as \texttt{MAUVE} \cite{pillutla2021mauve}, which approximate human-like text properties. On the other hand, \citet{garces-arias-etal-2024-adaptive} have observed that nucleus sampling could produce text with inconsistent semantics, although with respect to the prompt.
Further work, besides exploring longer completion windows (taking into account some of the above decoding issues), could also investigate the interpretative aspects, by which semantic relations within an isotopy are also driven by inference, something initially supported by the role of CoT in our benchmark performance, which decreased marginally (from 69.4\% to 61\%) when removing the CoT step. 
Beyond its contribution to the linguistic debate surrounding LLM, our work may also have practical implications, deriving from the potential use of isotopies to study generated text cohesion, or to underpin feedback supporting new forms of instruction-based training. 

\bibliography{Iso_Bib}
\bibliographystyle{acl_natbib}

\end{document}